\documentclass[letterpaper, 10 pt, conference]{ieeeconf}  

\IEEEoverridecommandlockouts                              

\usepackage{array,multirow,graphicx} 
\usepackage{epsfig} 
\usepackage{mathptmx} 
\usepackage{times} 
\usepackage{amsmath} 
\usepackage{amssymb}  
\usepackage{gensymb}
\usepackage{float}
\usepackage{xcolor}
\usepackage[normalem]{ulem}
\usepackage{wrapfig}
\definecolor{yellow}{rgb}{0.0, 0, 0.0}
\newenvironment{thisnote}{\par\color{yellow}}{\par}

\title{\LARGE \bf
Multi-modal Experts Network for Autonomous Driving
}

\author{Shihong Fang$^{*}$ and Anna Choromanska$^{*}$
\thanks{$^{*}$The authors are with the Machine Learning Lab, Department of Electrical and Computer Engineering,
        New York University, 5 MetroTech Center, USA.
        {\tt\small \{sf2584, ac5455\}@nyu.edu}}%
}

\begin{document}
\maketitle
\thispagestyle{empty}
\pagestyle{empty}

\begin{abstract}

End-to-end learning from sensory data has shown promising results in autonomous driving. While employing many sensors enhances world perception and should lead to more robust and reliable behavior of autonomous vehicles, it is challenging to train and deploy such network and at least two problems are encountered in the considered setting. The first one is the increase of computational complexity with the number of sensing devices. The other is the phenomena of network overfitting to the simplest and most informative input. We address both challenges with a novel, carefully tailored multi-modal experts network architecture and propose a multi-stage training procedure. The network contains a gating mechanism, which selects the most relevant input at each inference time step using a mixed discrete-continuous policy. The gating network chooses the camera input in a discrete way from among several mutually-exclusive sensors. Alternatively, the network chooses the LiDAR sensor, which covers the same field of view as the camera sensors, and identifies continuously in real-time the part of its depth map with a narrow field of view that is useful for steering autonomously. We demonstrate the plausibility of the proposed approach on our 1/6 scale truck equipped with three cameras and one LiDAR. 

\end{abstract}


\vspace{-0.17in}
\section{Introduction}
	
The advancements in deep learning~\cite{doi:10.1162/neco.1989.1.4.541} enabled artificial neural networks to invade a diverse set of real-life applications that need to rely on systems capable of learning, among which autonomous driving is an important and still largely unsolved task. Using multiple sensors improves perception abilities, leads to a better understanding of the environment, and consequently increases the safety and reliability of deep-learning-based autonomous platforms (some examples in the context of self-driving cars include~\cite{huval2015empirical,DBLP:journals/corr/BojarskiTDFFGJM16, chen2015deepdriving}). Nevertheless, it is highly non-trivial to teach the autonomous vehicle to use the information registered by different sensors potentially capturing various modalities. When there exists an overlap of information as is the case for camera and LiDAR registrations that cover the same or partially overlapping field of view, one needs to ensure the network is evenly trained to use any of the available relevant inputs. Otherwise, the failure of the sensor that the network was preferring at training leads to significant performance degradation. This could be avoided if instead of ignoring the information provided by other relevant sensors, the network would adapt to them as well. Another obstacle when using many sensors is the explosion of computational complexity. Addition of an extra sensor requires the expansion of the network to accommodate an extra feature extractor and increases computational cost.
This largely increases the inference time step and eventually paralyzes vehicle's real-time operation. The objective of this work is to address the above challenges. 

We propose a network that contains a feature extractor per each sensor and a gating mechanism that activates only a subset of feature extractors at each inference time step. The key component of the proposed architecture is the gating mechanism, which selects the most informative input in real time under strict computational constraints. It therefore efficiently switches the attention of the network among inputs (in other words experts).  In our setting, we have three camera inputs that are mutually exclusive, i.e. they have non-overlapping field of views (each of them covers $60$ degrees), and one LiDAR sensor that covers $180$ degrees\footnote{The LiDAR covers $360$ degrees, however we use only the front-facing part ($180$ degrees) as we only consider driving forward.} and therefore has an overlapping field of view with the cameras. We consider two schemes of input selection. The first one is the discrete one and occurs when selecting the proper sensor. The second one is the continuous one and occurs when selecting a relevant part of the information delivered by a single sensor. This is done for LiDAR, which has a very wide field of view. The proposed concept naturally provides a defensive mechanism against sensor failures as those will be considered as irrelevant ones and will not contribute to the final prediction. Moreover, wasteful computations correlated with extracting features from sensory inputs irrelevant or redundant for driving are avoided.

We also propose a multi-stage training procedure and show that it is superior compared to the other end-to-end learning techniques. We show that the end-to-end scheme does not result in the network that has the ability to select most relevant sensors. Note that the small gating network has its own feature extractors which should be correlated with the main feature extractors. The reason for the failure of the end-to-end approach is that it does not learn this correlation. 

We empirically verify our approach on our autonomous platform equipped with three cameras and LiDAR and provide evidence that our network correctly selects the proper sensor and works well in the discrete and continuous selection regimes, robustly predicts the vehicle's steering command, and requires significantly fewer computations than common baselines.

This paper is organized as follows: Section~\ref{sec:rw} provides a literature review, Section~\ref{sec:network} describes our approach, Section~\ref{sec:exp} demonstrates experimental results, and Section~\ref{sec:con} concludes the paper.


\section{Related Work}
\label{sec:rw}
\vspace{-0.035in}
This paper considers an end-to-end imitation learning framework for steering autonomous driving platforms that take raw sensor inputs and predict the steering command. \textcolor{yellow}{We consider deep learning framework since it allows for the automatic extraction of data features that are furthermore useful for a particular learning task and potentially more scalable to real-life systems. Alternative non-deep learning techniques for multi-sensor data fusion for autonomous driving were investigated in~\cite{maddern2016real,de2018fusion,lahat2015multimodal,luo2002multisensor}.} The functional mapping between inputs and outputs in the end-to-end setting was realized in past works using fully-connected networks trained on synthetic data~\cite{pomerleau1989alvinn} or convolutional neural networks (CNNs) trained on real recorded data without~\cite{lecun1998gradient,muller2006off, xu2017end} and with~\cite{DBLP:journals/corr/BojarskiTDFFGJM16, muller2006off} augmentation. All of the above mentioned works rely only on a single camera sensor at inference. The conditional imitation learning approach~\cite{Codevilla2018} relies on a single camera and uses camera images and navigation commands  to steer the vehicle through road intersections in desired directions. This work was further extended~\cite{amini2018variational} to use both camera images and maps to generate a probability distribution over possible steering commands. Finally,~\cite{Fang2019ReconfigurableNF} proposes a reconfigurable network that uses images captured from three front-facing cameras as inputs to navigate the autonomous driving car in an indoor environment. There are only a few works proposing end-to-end learning frameworks accommodating multiple sensing modalities such as LiDARs and cameras~\cite{pfeiffer2017perception,patel2017sensor,chen2017multi, Liu2017LearningEM}. Our work concerns precisely such setting and engages the gating mechanism for relevant sensor selection unlike existing approaches.

The gating mechanism was explored in the literature in various settings and applications. The first system incorporating this mechanism~\cite{jacobs1991adaptive} was the modular version of a multi-layer supervised network, where many separate networks (experts) were operating on the same input but they were handling different learning sub-tasks. 
The gating mechanism has often been used as an enabling technique for scaling algorithms to large data sets and has been flexible enough to adapt to different learning systems, including those with experts realized as SVMs~\cite{collobert2002parallel} or neural networks~\cite{eigen2013learning,bengio2015conditional,bolukbasi2017adaptive,shazeer2017outrageously} and those where experts were hierarchically-structured~\cite{jordan1994hierarchical}. The works on conditional computation~\cite{eigen2013learning,bengio2015conditional, bolukbasi2017adaptive} explore the notion that irrelevant or redundant experts need not be activated at all when forming a prediction. Sparsely-gated mixture-of-experts~\cite{shazeer2017outrageously} realizes an extreme version of this concept, where a large number of experts is used, but only a very small subset is activated at each inference time step. None of the gating schemes discussed above was applied to autonomous driving, but they were explored in the context of more traditional AI problems such as language modeling, machine translation, or image classification. Our work falls into the category of conditional computations schemes, but as opposed to existing methods experts do not share the same input. Finally, gating mechanism was also used for sensor fusion~\cite{mees2016choosing,patel2017sensor} in autonomous driving, where all experts are active. These approaches therefore are not concerned with constraints on the computational budget since they allow wasteful feature extractions from irrelevant or redundant sensors.  Our work fundamentally differs from these works as in our case such experts are identified at each inference time step and are deactivated, which leads to computational savings. 
	

\section{Multi-modal Experts Network}
\label{sec:network}
\subsection{Network architecture}

\begin{figure}[ht]
\vspace{-0.40in}
    \centering
    \includegraphics[width=\columnwidth]{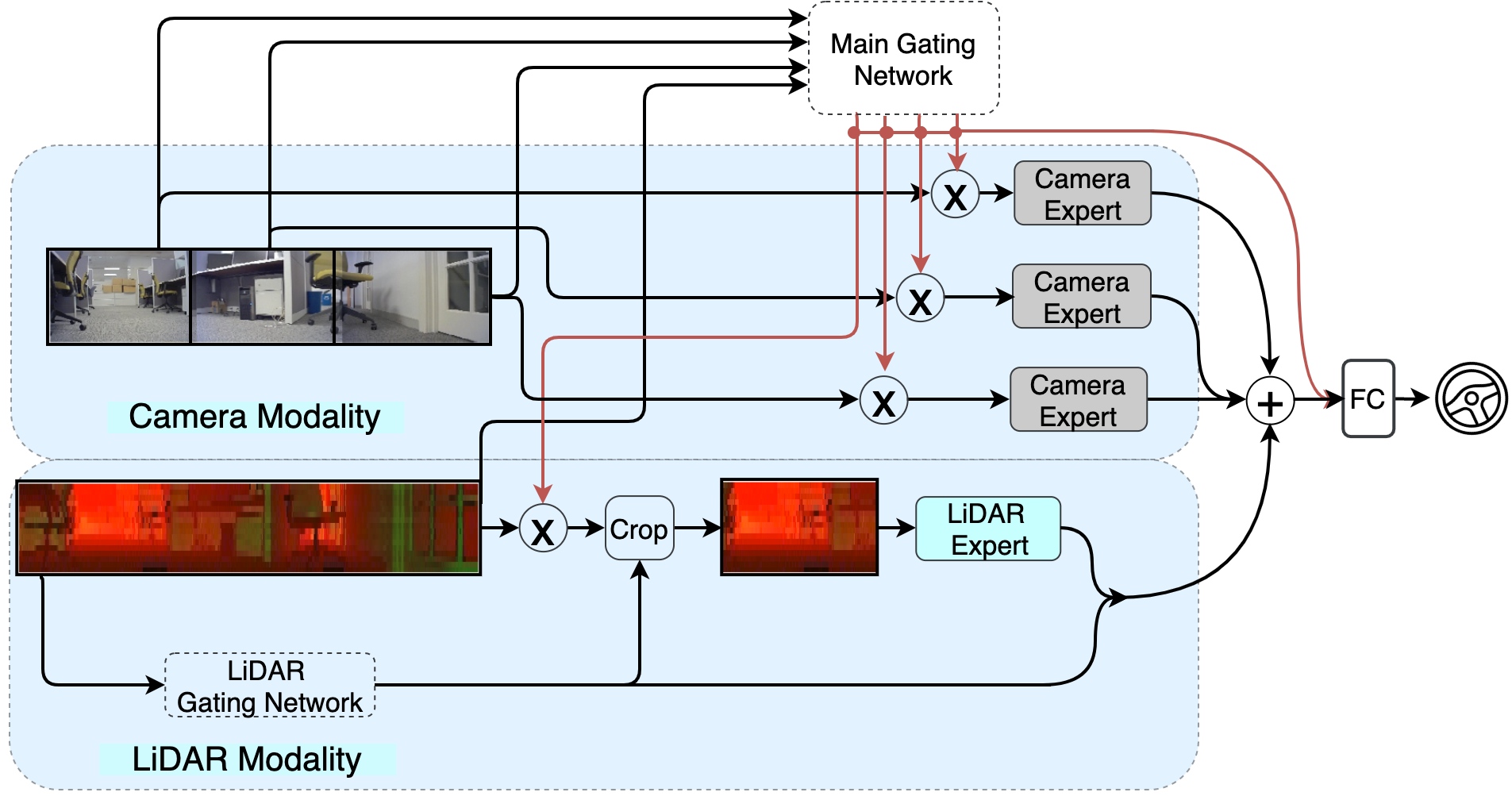}
    \caption{The architecture of the proposed network. There are two highlighted sections: Camera Modality at the top, and LiDAR Modality at the bottom. In the Camera Modality, there are three inputs, each from a different camera, followed by the corresponding feature extractors (Camera Experts). In the LiDAR Modality, the single input from the LiDAR is used. All the camera inputs and \textcolor{yellow}{the entire} LiDAR input are passed to the Main Gating Network that decides which input contains the most relevant information for driving at the moment and activates the corresponding expert. If LiDAR is selected, its input is first processed by the LiDAR Gating Network that decides which part of the input will be cropped and used for further processing. The cropped LiDAR input is passed to the LiDAR feature extractor (expert). Finally, the output of the LiDAR Gating Network is concatenated with the output of the LiDAR expert. The output of the Main Gating Network is concatenated to the output of the active expert and passed to the final Fully-Connected Network (FC) that forms the prediction of the steering command.}
    \label{fig:arch-final}
    \vspace{-0.15in}
\end{figure}

The proposed multi-modal experts network is illustrated in Fig.~\ref{fig:arch-final}. The network is designed to handle inputs coming from three cameras and one LiDAR. Thus the network has four inputs in total coming from two sensory modalities. We will refer to those inputs as $\mathbf{x}=[x_1,x_2,x_3,x_4]$, where $x_1,x_2,x_3$ are the camera inputs and $x_4$ is the LiDAR input. The final prediction $\hat{y}$ of the multi-modal experts network can be therefore expressed in the following way:

\vspace{-0.2in}

\begin{align}
    g^M_i &= G^{M}(\mathbf{x})_i \textcolor{yellow}{, i=1, 2, 3, 4} \label{eq:0} \\
    v_i &= E_i(x_i), i=1, 2, 3 \label{eq:2}\\
    x^{\prime}_4 &= G^L(x_4) \odot x_4, \label{eq:1}\\
    v_4 &= \left[E_4(x^{\prime}_4), G^L(x_4)\right], \label{eq:3}\\
    \hat{y} &= F\left(\left[\sum_{i=1}^{4} g^M_i v_i,
    \; G^{M}(\mathbf{x})\right]\right), \label{eq:4}
\end{align}

\noindent In the first equation, $G^{M}(\mathbf{x})_i$ denotes $i^{\text{th}}$ output of the Main Gating Network, where at each time instant, all $G^{M}(\mathbf{x})_i$s are equal to $0$ except one, which is equal to $1$. Therefore, $g^M_i$ encodes the information of the selection of the Main Gating Network, which can been seen as an indicator showing which sensor carries the most relevant information for driving at any given moment. \textcolor{yellow}{In our setting, the data captured by the sensors are spatially redundant with each other. In order to minimize the amount of required computations to steer a vehicle, we  assume that only one expert, that is chosen by the Main Gating Network, will actually be activated.} The extension to the case where multiple sensors are used is easy to achieve. Concretely, if we want to keep $k$ experts, the output of the Main Gating Network should be modified accordingly, e.g. if one wants to use 2 out of 4 sensors and use weights for each input, one should zero-out the smallest and the second smallest weights. $E_1, E_2, E_3$ are feature extractors of camera images (Camera Experts) and $E_4$ is the feature extractor of LiDAR image (LiDAR Expert). 
If one of the cameras is selected, we can get feature vector $v_i$ from the output of the Camera Expert in Equation \ref{eq:2}. Equation \ref{eq:1} and \ref{eq:3} show the process of obtaining LiDAR feature vector when the LiDAR expert is picked.
In Equation \ref{eq:1}, $G^L$ presents the LiDAR Gating Network and its output $G^{L}(\cdot) \in [-1,1]$ such that $-1$ corresponds to the leftmost $60$ degree LiDAR image, $1$ corresponds to the rightmost $60$ degree LiDAR image, and internal values of the $[-1,1]$ interval correspond to the proportionally shifted $60$-degree image window. The $\odot$ denotes the cropping operation and the selected $60$ degree image $x^{\prime}_4$ is used for further processing. In Equation \ref{eq:3}, the output of the LiDAR gating network is concatenated with the LiDAR feature vector $E_4(x_4^{\prime})$ to form the LiDAR feature vector $v_4$. Finally, in Equation \ref{eq:4}, the resulting feature vector is formed by concatenating the feature vector coming out from the chosen expert and the output of the Main Gating Network.
$F$ is the action of the final fully-connected network \textcolor{yellow}{and transforms the features into the final prediction label.}
This construction ensures that the resulting feature vector encodes the information extracted from the selected input as well as the identity of the selected sensor, and in case this sensor is LiDAR also the information about the part of the input that was used. 

\subsection{Training procedure}
We propose a carefully tailored procedure for training the multi-modal experts network. The training procedure is designed to enable efficient and robust learning of the behavior of the gating mechanism from the training data, which do not carry any information about the identity of the most relevant sensor at any given moment in time. \textcolor{yellow}{The network training procedure that we propose consists of three main steps. In the \textbf{\textit{first step}}, we train the network equipped with the gating mechanism to predict the steering command based on the LiDAR input. Thus the network effectively learns which part of the LiDAR input is relevant (continuous sensor selection). In the \textbf{\textit{second step}}, we train another network, also equipped with the gating mechanism, to predict the steering command from either one of the cameras or LiDAR input. Thus the network effectively learns to choose between sensors and modalities (discrete sensor selection). In the \textbf{\textit{third step}}, we incorporate the LiDAR network trained in the first step into the multi-modal network trained in the second step and fine-tune everything to simultaneously  choose the proper sensor (learned in step 2) and the relevant part of the LiDAR input (learned in step 1), if its modality was chosen.}

\textcolor{yellow}{Moreover, for the purpose of training the gating networks (either the LiDAR Gating Network from the first step or the Main Gating Network from the second step) realizing the gating mechanism, we consider two sub-steps: 
\begin{itemize}
    \item We train the entire network (all feature extractors and the gating mechanism that uses those extractors) to predict the steering command. The by-products of this training are the labels for the gating network that indicate relevant sensors for driving. Since the gating network in this step uses the same feature extractors as the main network it is able to learn sensor relevance without explicit labels. (Point 1.1 explains this for the LiDAR Gating Network and point 2.1 explains this for the Main Gating Network.)
    \item We train a separate computationally-constrained architecture of the gating network (thus we effectively reduce the size of the gating network) using labels that were inferred in the previous stage. (Point 1.2 explains this for the LiDAR Gating Network and point 2.2 explains this for the Main Gating Network.)
    \end{itemize}
We next describe the training procedure in details.}

\subsubsection{\textbf{First Step}}
\begin{figure}[t]
\vspace{0.1in}
    \centering
    \includegraphics[width=0.85\columnwidth]{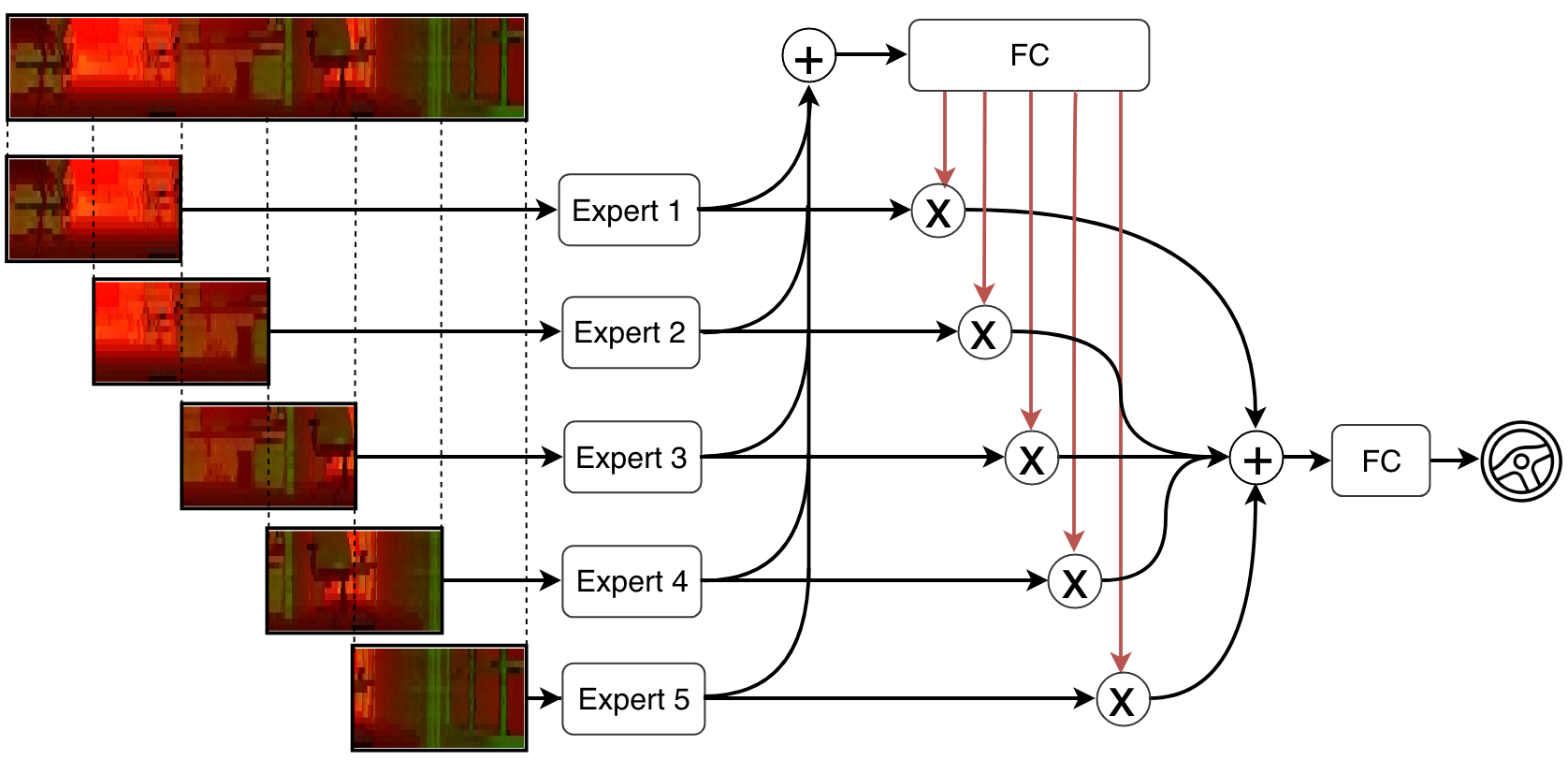}
    \caption{The architecture of the network used in the \textcolor{yellow}{1.1} step of the training procedure. There are $5$ experts: leftmost, left, center, right, and rightmost. 
    }
    \vspace{-0.2in}
    \label{fig:arch-5lidar}
\end{figure}
\textcolor{yellow}{\textit{Continuous sensor selection}}

\textcolor{yellow}{\textit{1.1 Obtaining labels for the gating mechanism}}

We train the network with the architecture given in Fig.~\ref{fig:arch-5lidar}. Its input is formed by splitting the LiDAR input into five partially overlapping segments (they overlap one-half), each covering $60$ degree field of view. Each segment is processed by the corresponding expert. The feature vectors obtained from the experts are processed by the fully-connected network that realizes a function of the gating mechanism. Therefore, the experts and the mentioned fully-connected network form a gating network. Finally, the feature vectors are scaled by the outputs of the gating network and passed to another fully-connected network that predicts the steering command. We train this entire model in an end-to-end fashion. We designed a cost function comprising three carefully selected terms to ensure desired performance of the gating behavior. The first term is the prediction loss measuring the discrepancy between the true and predicted steering command and to ensure the final output of the network can successfully steer the robot.
\vspace{-0.1in}
$$\mathcal{L}_{prediction} = \left\lVert y - \hat{y}\right\rVert^2_2.$$

\noindent To ensure that the output of the gating network is sparse so that the most informative input can easily stand out, we introduced the second term that we refer to as sparsity loss:
\vspace{-0.15in}
$$\mathcal{L}_{sparsity} = \sum_i -g_i \log(g_i),$$

\noindent where $g_i$s are the outputs of the gating network. The outputs of  the  gating  network  are  obtained  after a softmax layer,
\textcolor{yellow}{where $g_i = \frac{\exp{(o_i)}}{\sum^n_{i=1}\exp{(o_i)}}$($o_i$ is the input to the softmax layer)}, thus $\sum_ig_i = 1$. In this stage we allow more than one $g_i$ to be non-zero. The final term of the loss is the negative entropy computed with respect to the distribution of the gating network outputs. It is given as
$$\mathcal{L}_{nentropy} = \sum_i p_i  \log(p_i),$$
where  $p_i = \frac{1}{B}\sum_{j=1}^B g_{i,j}$, where $g_{i,j}$ is the $i^{th}$ output of $j^{th}$ gating network observation in one batch and $B$ is the training batch size. Minimizing this term prevents the output of the gating network from collapsing and favoring a small and fixed subset of experts over the entire data batch, while zeroing out the rest. 

The objective function has therefore the following form:
\begin{align}
    \mathcal{L} = \mathcal{L}_{prediction} +\alpha \mathcal{L}_{sparsity} + \beta \mathcal{L}_{nentropy},
\end{align}
where $\alpha$ and $\beta$ are the hyper-parameters controlling the importance of sparsity and negative entropy terms. We aim at minimizing this objective. In the result, we obtain the trained  gating network that we then use for generating labels for the desired gating mechanism.

\textcolor{yellow}{\textit{1.2 Training the LiDAR Gating Network}}

We train a separate LiDAR Gating Network that using knowledge distillation \cite{hinton2015distilling}. This one predicts which part of the LiDAR input is the most relevant for driving. We train the LiDAR Gating Network using the gating mechanism obtained in the previous training step as the teacher network. Specifically, the labels $g_L$ for training the new gating mechanism are calculated as $g_L = \sum_i g^L_i r_i$, where $g^L_i$ are outputs of the reference gating mechanism obtained in the first step of the training procedure and $r_i=[-1.0,-0.5,0.0,0.5,1.0]$ are the values of the $g_L$ corresponding to the segments of the LiDAR image described before.

\textcolor{yellow}{\textit{1.3 Fine-tuning}}

In the third sub-step of the training procedure, we train the network with the architecture depicted in Fig.~\ref{fig:arch-lidar}. Here we are using the LiDAR Gating Network obtained in the previous step of training and we keep its weights fixed. The network uses LiDAR input which first is processed by the LiDAR Gating Network. Based on the LiDAR Gating Network output, part of the input is cropped. In particular, we crop a segment covering $60$ degree field of view. The segment range is calculated as
$c_l = 60^{o}g_L - 30^{o}; c_r = 60^{o}g_L + 30^{o}$, where $c_l$ and $c_r$ are the left and right edges of the cropped segment respectively and $g_L$ is the output of the LiDAR Gating Network that is in the range $[-1, 1]$. Next, the cropped segment is processed by the feature extractor (expert). The output of the LiDAR Gating Network and the obtained feature vector are concatenated together and fed to the fully-connected layer, which produces the steering command. At the end of this step, we obtain a functional network with the single LiDAR input for the autonomous driving \textcolor{yellow}{and we refer to this network as ``LiDAR with gating'' network.}
\subsubsection{\textbf{Second Step}}
\begin{figure}[t]
    \vspace{0.1in}
    \centering
    \includegraphics[width=0.95\columnwidth]{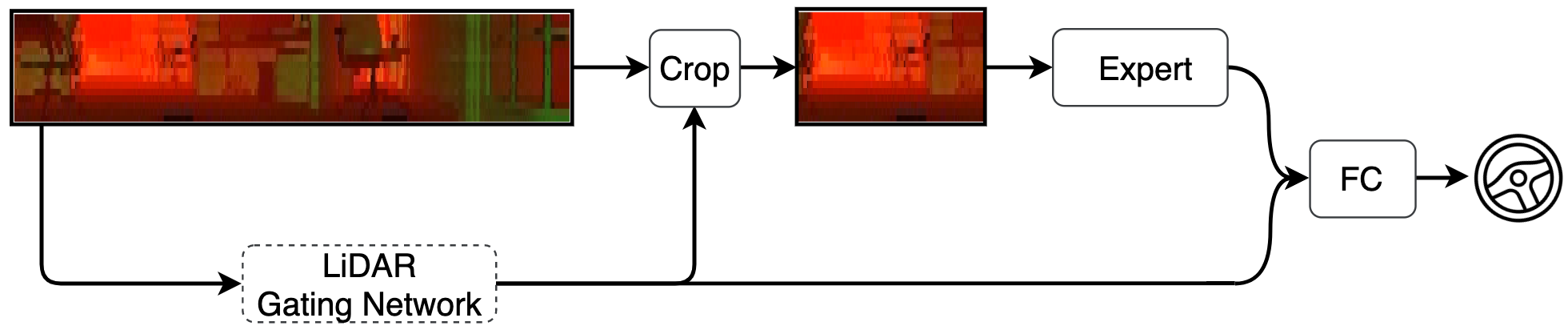}
    \caption{The architecture of the neural network used in the \textcolor{yellow}{1.3} step of the training procedure. \textcolor{yellow}{The dashed box indicates the LiDAR Gating Net is fixed during training.}
    }
    \label{fig:arch-lidar}
    \centering
    \includegraphics[width=\columnwidth]{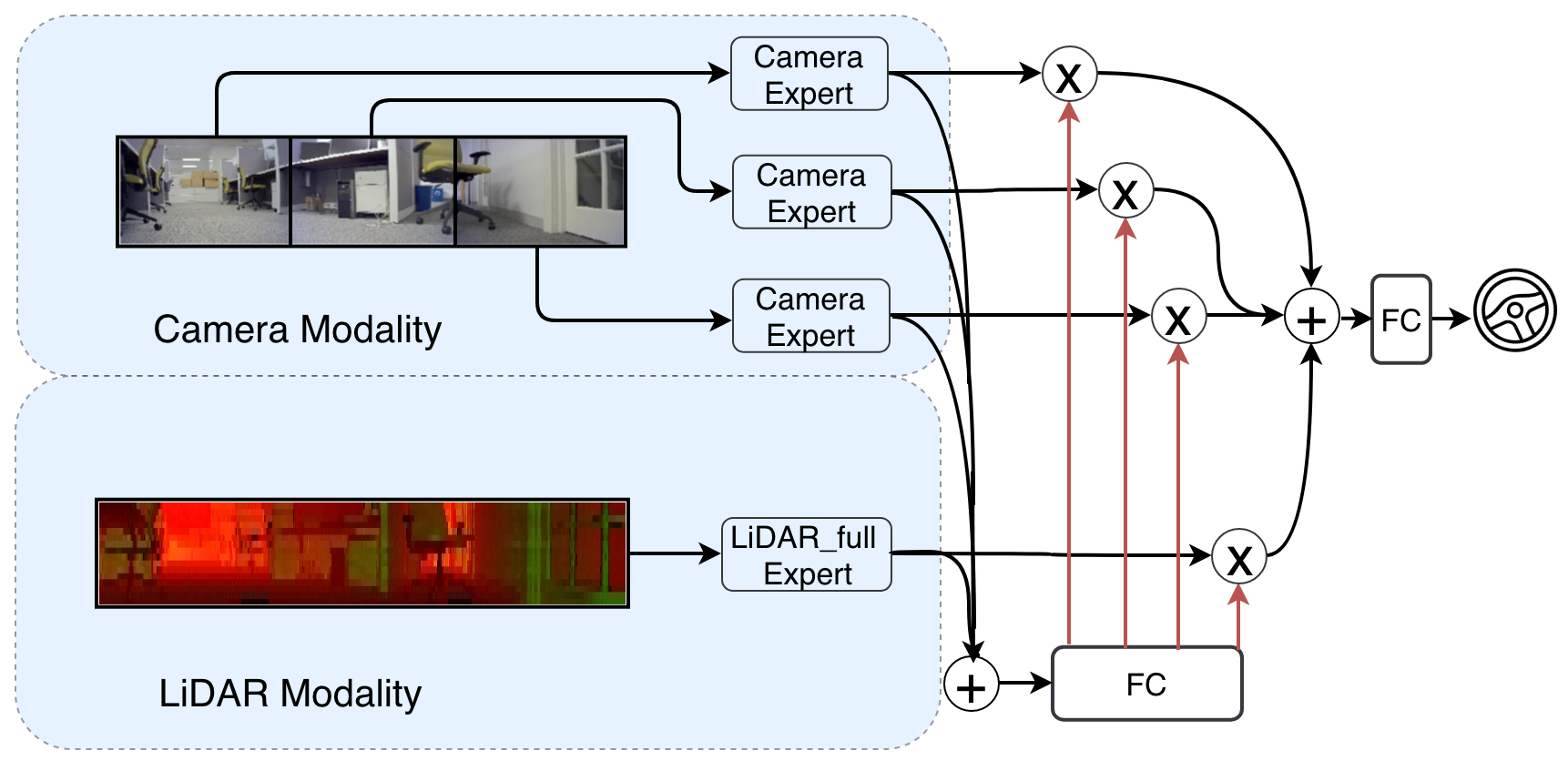}
    \caption{The architecture of the neural network used in the \textcolor{yellow}{2.1} step of the training procedure. 
    In the Camera Modality, there are three inputs, each from different camera, followed by the corresponding feature extractors (Camera Experts). In the LiDAR Modality, the single input from the LiDAR is used and processed by feature extractor of the entire LiDAR image (LiDAR\_full Expert). 
    }
    \label{fig:arch-4sensor}
    \vspace{-0.2in}
\end{figure}
\textcolor{yellow}{\textit{Discrete Sensor Selection}}

\textcolor{yellow}{\textit{2.1 Obtaining labels for the gating mechanism}}

First, we train the network with the architecture depicted in Fig.~\ref{fig:arch-4sensor}. The architecture uses three inputs from camera modality and one input from LiDAR modality and is capable of handling sensor fusion. The feature vectors obtained from three camera experts and one LiDAR\_full expert (feature extractor of the entire LiDAR input) are also used by the gating network as feature extractors, thus it can be seen that the additional fully connected layers realize the functionality that we call the gating mechanism (selecting relevant inputs). Finally, the feature vectors are scaled by the outputs of the gating network and passed to another fully-connected network that predicts the steering command. We train this entire model in an end-to-end fashion using the same objective function as the one introduced in the 1.1 sub-step.

\textcolor{yellow}{\textit{2.2 Training the Main Gating Network}}

As a result of the 2.1 step of training, we obtain the trained gating network that serves as a reference delivering labels for training the Main Gating Network from Fig.~\ref{fig:arch-final}. The latter network realizes the gating mechanism showing which input (one of three camera inputs or LiDAR input) is most relevant for driving. Note that we used the same training strategy in the 1.2 step of training. To be more specific, the labels $g^{M}_i$ for training the Main Gating Network are calculated as
\vspace{-0.10in}
\begin{align}
    j &= \arg\max g^M_i, \\
    g^{M}_j &= 1, \\
    \bigwedge_{i \neq j}g^{M}_i &= 0,
    \end{align}
\noindent where $g^M_i$ are the outputs of the reference gating mechanism obtained in the previous step of the training procedure. In this case, the Main Gating Network is trained to select one, the most relevant input for driving.

\subsubsection{\textbf{Third Step}}
\textcolor{yellow}{\textit{Training the Multi-Modal Experts Network}}

In the final step of the training procedure, we train our proposed multi-modal experts network depicted in Fig.~\ref{fig:arch-final}. In LiDAR modality, we use part of the network architecture in the \textcolor{yellow}{first} step but without final fully-connected layer to process LiDAR information. In this way, if the LiDAR expert is selected, the computation can be further optimized. Here we keep the weights of the LiDAR Gating Network obtained in the \textcolor{yellow}{first} step and the Main Gating Network obtained in the \textcolor{yellow}{second} step fixed. Thus, in this step, we train all the experts and the final fully-connected layer that produces the steering command. In the result, we obtain the final network that identifies the most relevant input for predicting the steering command and activates only the corresponding expert at each inference time step.

\section{Experiments} 
\label{sec:exp}
\subsection{Hardware Overview}
\begin{wrapfigure}{l}{0.5\columnwidth}
    \vspace{-0.2in}
    \centering
    \includegraphics[width=0.5\columnwidth]{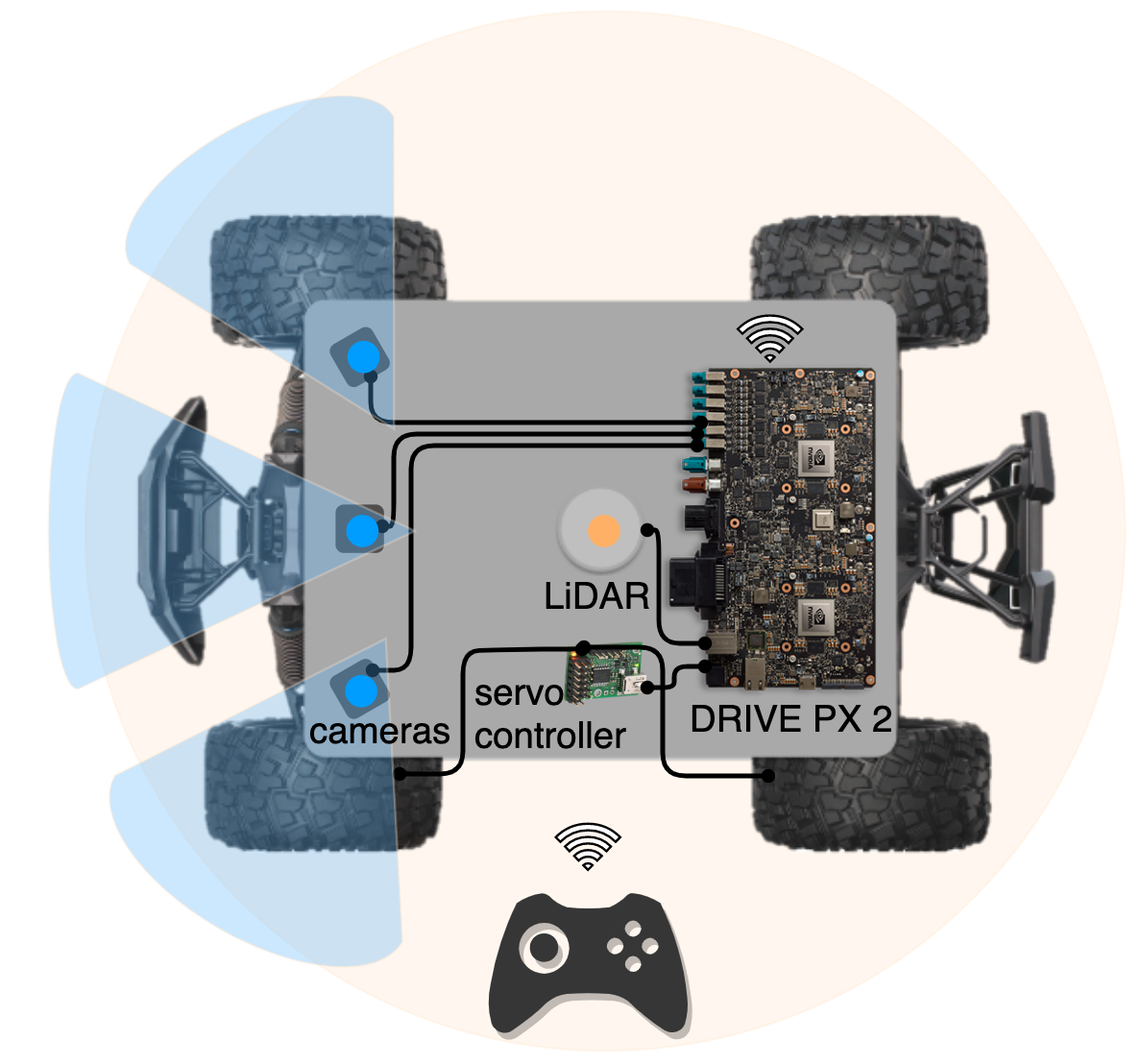}
    \caption{The block diagram of the autonomous platform used in the experiments.}
    \vspace{-0.2in}
    \label{fig:hardware}
\end{wrapfigure}
The block diagram of the autonomous platform that we used for data collection and system testing is shown in Fig.~\ref{fig:hardware}. The platform is based on the Traxxas X-Maxx remote control truck and equipped with Drive PX2 that performs computations. We incorporated the following sensors on the platform:

\begin{itemize}
    \item three SEKONIX AR0231 GMSL cameras that are facing the front of the platform and covering non-overlapping views. Each camera has $60$ degrees horizontal field of view. The center camera is oriented straight. The side cameras are mounted at angles to allow capturing front side views. 
    \item single Velodyne VLP-16 LiDAR with $16$ lasers that cover a $30$ degree vertical field of view and $360$ degree horizontal field of view.
\end{itemize}
   
Controlling the actuators of the autonomous platform was done using Micro Maestro 6-Channel USB Servo Controller. When collecting the data, the wireless game-pad was used to drive the platform. The platform was controlled by an experienced human driver to ensure it drives in the center of the path \textcolor{yellow}{without hitting the walls of the corridors or obstacles}. The captured steering commands were used as ground truth for training and evaluation. The steering commands together with the corresponding cameras and LiDAR inputs captured by the sensors were saved on the Drive PX2 local storage. At testing, the platform runs in real time and is steered with the commands predicted by \textcolor{yellow}{the} deep learning gating-based system. The speed is dictated by the operator of the platform.

\subsection{Data Pre-processing}

The SEKONIX AR0231 GMSL cameras have resolution $1920\times1208$. We scale down the images captured by the cameras to the size $192\times120$. We use three channels (RGB) for each camera. 

The LiDAR captures data in the form of a point cloud, which we convert into a depth map. We utilize only the front-facing $180$ degrees field of view of the LiDAR input as the car is only required to drive forward. The resulting depth map has size $450\times16$. We use two channels, one for distance and one for reflectivity. The exemplary images from cameras and LiDAR are shown in Fig.~\ref{fig:arch-final} in camera and LiDAR modality respectively.


The values of recorded steering commands are normalized to the range [-1,1], where $-1,0,1$ corresponds to driving maximally left, straight, and maximally right respectively.
The data are time-synchronized by dropping redundant frames and are recorded at a rate of 5 examples per second.
 We collected $43,812$ training examples in total. Each sample is represented by three images from the cameras, one depth map from LiDAR, and the corresponding steering command. We group the steering commands of all training examples $\textbf{X}$ into 7 categories $\textbf{X}_{1:7}$: \textcolor{yellow}{$|\textbf{X}_{1}|=7297$ for $y \in [-1,-0.67)$; $|\textbf{X}_{2}|=2094$ for $y \in [-0.67,-0.33)$; $|\textbf{X}_{3}|=9706$ for $y \in [-0.33,0.00)$; $|\textbf{X}_{4}|=15127$ for $y=0.00$; $|\textbf{X}_{5}|=1555$ for $y \in (0.00, 0.33]$; $|\textbf{X}_{6}|=2915$ for $y \in (0.33,0.67]$; $|\textbf{X}_{7}|=5118$ for $y \in (0.67,1]$}. In each training epoch, we sample $10,000$ examples uniformly at random from each category to balance the data set. We also collected the test data containing $2,384$ examples. The train and test data sets were collected in different parts of the same building. \textcolor{yellow}{All reported results in the next sections were obtained on the test data set.} 
 
\subsection{Training Details and Evaluation}

\begin{figure}[t]
    \centering
        \vspace{0.13in}
    \includegraphics[width=\columnwidth]{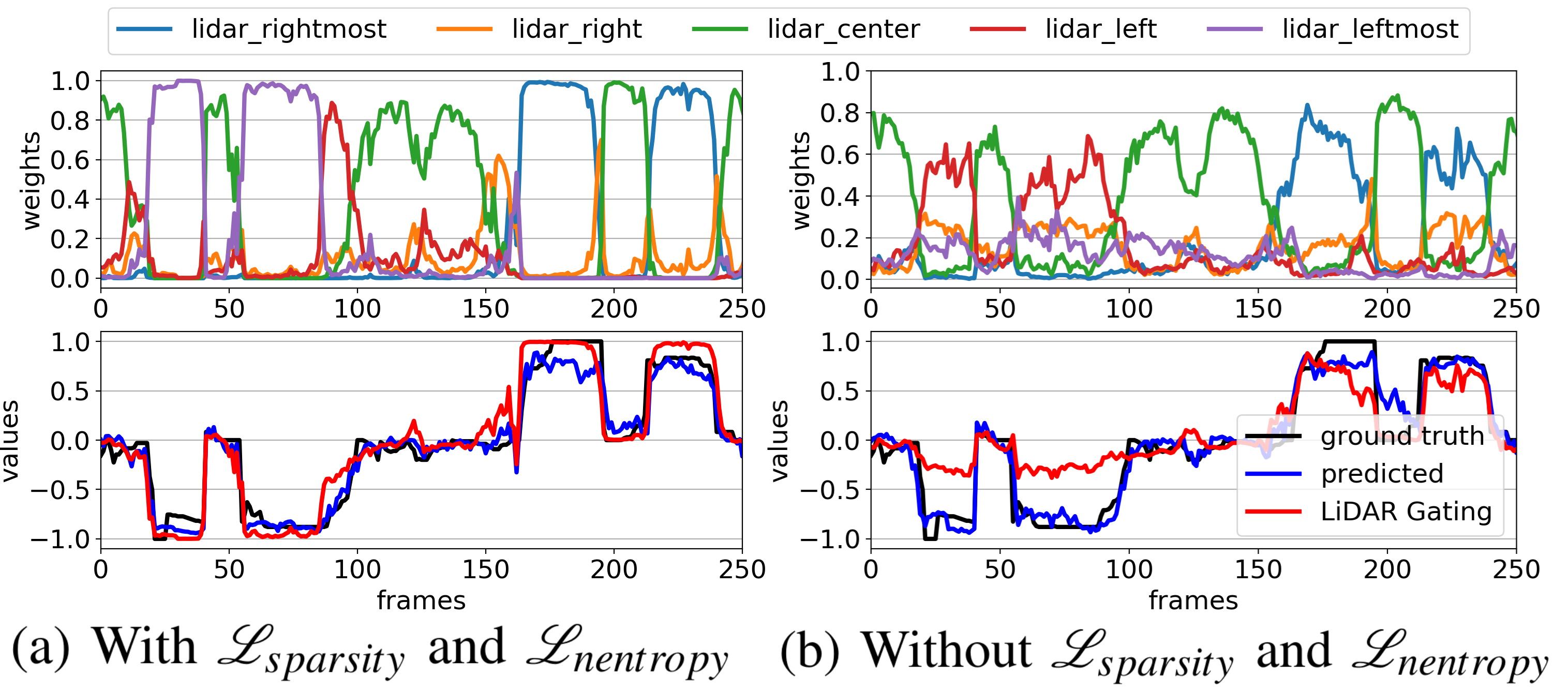}
    \caption{\textbf{Top:} The output of the gating network obtained from the \textcolor{yellow}{1.1} step of the training procedure indicating the relevance of each expert for predicting the steering command. \textcolor{yellow}{The corresponding network architecture is shown in Fig.~\ref{fig:arch-5lidar}.} \textbf{Bottom:} The comparison between the actual and predicted steering command. The network \textcolor{yellow}{(shown in Fig.~\ref{fig:arch-lidar})} obtained from the \textcolor{yellow}{1.3} step of the training procedure is used for prediction. In red we show the output of the LiDAR Gating Network. \textcolor{yellow}{Only the first 250 test frames are shown here due to the space limitation.}}
    \label{fig:lidar-gating}
    \vspace{-0.23in}
\end{figure}

We use ADAM optimizer in all steps of the training procedure. Hyperparameters are tuned by grid search to ensure best performance. In the \textcolor{yellow}{1.1} step, we use the learning rate equal to $0.001$ and the following hyper-parameters for the objective function: $\alpha=0.001, \beta=0.0016$. The performance of the gating mechanism obtained in the \textcolor{yellow}{1.1} step of the training procedure is reported in Fig.~\ref{fig:lidar-gating}(a)(top). In the \textcolor{yellow}{1.2} step, we train the LiDAR Gating Network using knowledge distillation. The loss terms consist of distillation loss and student loss. The distillation loss is measured as the KL divergence loss between soft labels of the teacher network and soft predictions of the student model. The student loss is defined as the mean square error (MSE) between the predicted labels and the labels produced by the reference gating mechanism. In this step, the learning rate was set to $0.001$, the distillation temperature is set to 2 and the weight on the distillation loss is 0.9. In the \textcolor{yellow}{1.3} step, the learning rate was set to $0.001$ and the \textcolor{yellow} {test} performance of the network is shown in Fig.~\ref{fig:lidar-gating}(a)(bottom). The network clearly can learn to predict the correct steering command and properly selects most relevant part of the LiDAR input at given time instant. 

\begin{figure}[t]
    \centering
    \includegraphics[width=\columnwidth]{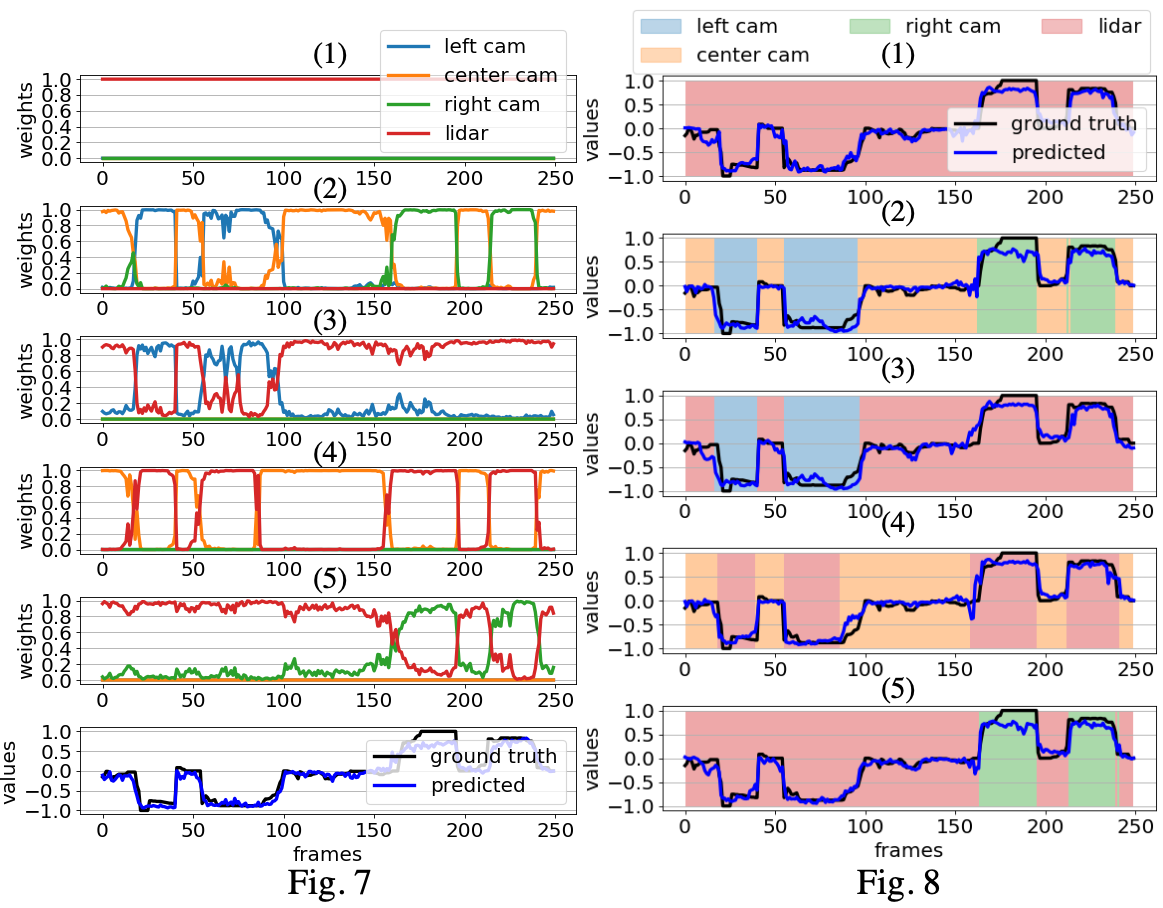}
    \vspace{-0.2in}
    \caption{\textbf{Top:} The outputs of the individual gating network obtained in the \textcolor{yellow}{2.1} step of training procedure when the inputs are (1)-(5). \textcolor{yellow}{The network architecture is shown in Fig.~\ref{fig:arch-4sensor}}.
    \textbf{Bottom:} The comparison between the actual and predicted steering commands for the $5^{\text{th}}$ network. \textcolor{yellow}{Results of the other networks are similar.}}
    \label{4s-final-result}
    \vspace{0.12in}
    \caption{The \textcolor{yellow}{test} performance of the final network \textcolor{yellow}{(shown in Fig.~\ref{fig:arch-final})} when the inputs are (1)-(5). The curves show the comparisons between the actual steering command and predicted steering command. The color-shaded areas indicate the selection of the Main Gating Network.}
    \vspace{-0.2in}
    \label{5-branch-result}
\end{figure}

In the \textcolor{yellow}{2.1} step of the training process, we use the learning rate of $0.0001$. The hyper-parameters of the objective function are $\alpha=0.002$ and $\beta = 0$ (this is because there is no shared information between any two different inputs). In this step, we assume that at most one sensor will encounter a sensor failure. Thus, we separately train different instances of the same network with architecture depicted in Fig.~\ref{fig:arch-4sensor}, where for each instance we disable different sensors. We consider the following cases:
(1) LiDAR is enabled and at most one of the camera inputs is disabled
(2) LiDAR is disabled and all the camera inputs are enabled
(3) Left part of LiDAR is disabled and all the camera inputs are enabled 
(4) Center part of LiDAR is disabled and all the camera inputs are enabled 
(5) Right part of LiDAR is disabled and all the camera inputs are enabled. The \textcolor{yellow}{test} performance of the resulting five networks is depicted in Fig.~\ref{5-branch-result}. Clearly, the gating network can successfully select the most relevant sensor. For example, when LiDAR sensor is enabled the gating network always selects the LiDAR expert as the best sensor for driving. Otherwise, when the LiDAR sensor is disabled, the gating network chooses the most relevant camera. \textcolor{yellow}{This is intuitive as LiDAR input is more informative compared with RGB camera image and the information from the side sensors will be more important when the car makes turns.}

In the \textcolor{yellow}{2.2} step of the training procedure, the distillation loss is the same and the student loss is the cross-entropy loss between the predicted labels and the labels produced by the reference gating mechanism. In this step we use the ADAM optimizer with the learning rate set to $0.001$. The distillation temperature is set to 4 and the weight on the distillation loss is 0.9. We train the Main Gating Network using all the five gating networks obtained in the previous step as reference. The performance of the obtained Main Gating Network is reported in Table~\ref{tab:test-result-gating}. The network achieves high performance accuracies of choosing the correct sensor.

\begin{table}
  \begin{minipage}[t]{.37\linewidth}
    \centering
    \vspace{0.08in}
    \caption{Performance of the Main Gating Network with different inputs disabled}
    \label{tab:test-result-gating}
    \begin{tabular}{|c|c|}
    \hline
    Case & Accuracy(\%) \\
    \hline
    \hline
    (1) &  100\\
    \hline
    (2) & 93.41\\
    \hline
    (3) & 97.69\\
    \hline
    (4) & 94.33\\
    \hline
    (5) & 96.98\\
    \hline 
    \end{tabular}
  \end{minipage}\hfill
  \hspace{-0.43in}
  \begin{minipage}[t]{.52\linewidth}
    \centering
    \vspace{0.08in}
\caption{Performance of several multi-modal networks with different inputs disabled}
    \label{tbl:4s-final-result} 
    \begin{tabular}{|c|c|} 
    \hline
    Case & Test MSE loss(Ours/\textcolor{yellow}{BL/}NG/SD)\\
    \hline
    \hline
    (1) &  0.020/ \textcolor{yellow}{0.026/} 0.023/ 0.023\\
    \hline
    (2) & 0.033/ \textcolor{yellow}{0.026/} 0.023/ 0.045\\
    \hline
    (3) & 0.026/ \textcolor{yellow}{0.026/} 0.022/ 0.038\\
    \hline
    (4) & 0.029/ \textcolor{yellow}{0.025/} 0.022/ 0.024\\
    \hline
    (5) & 0.024/ \textcolor{yellow}{0.025/} 0.022/ 0.026\\
    \hline 
    \hline 
    Avg.  & 0.026/ \textcolor{yellow}{0.026/} 0.022/ 0.031\\
    \hline 
    \end{tabular}
  \end{minipage}\hfill
  \mbox{}
\end{table}

\begin{table}[t]
    \centering
    \vspace{-0.15in}
    \caption{Performance and computational complexity comparison of different networks.(*indicates the multi-modal model)}
    \label{tbl:FLOPs} 
    {\color{yellow}\begin{tabular}{|l|c|c|c|}
\hline
Method                               & Network Name          & Test MSE loss                                        & FLOPs   \\ \hline
\multirow{6}{*}{\begin{tabular}[c]{@{}l@{}}End-to-end \\ training\end{tabular}} & LiDAR only            & 0.020                                                & 52.23M  \\ \cline{2-4} 
                                     & Single Camera         & 0.048                                                & 50.58M  \\ \cline{2-4} 
                                     & Three Cameras         & 0.036                                                & 151.48M \\ \cline{2-4} 
                                     & *Baseline(BL)          & see Table \ref{tbl:4s-final-result} & 204.69M \\ \cline{2-4} 
                                     & *NetGated(NG)          & see Table \ref{tbl:4s-final-result} & 204.90M \\ \cline{2-4} 
                                     & *Sensor Dropout(SD)    & see Table \ref{tbl:4s-final-result} & 204.69M \\ \hline
\multirow{5}{*}{\begin{tabular}[c]{@{}l@{}}Multi-step \\ training\end{tabular}} & LiDAR with gating     & 0.023                                                & 28.15M  \\ \cline{2-4} 
                                     & *Ours                  & see Table \ref{tbl:4s-final-result} & 35.71M  \\
                                     & chosen sensor: LiDAR  &                                                      &         \\ \cline{2-4} 
                                     & *Ours                  & see Table \ref{tbl:4s-final-result} & 58.15M  \\
                                     & chosen sensor: camera &                                                      &         \\ \hline
\end{tabular}}
    \vspace{-0.3in}
\end{table}

\begin{figure}[t]
    \centering
    \vspace{0.08in}
    \includegraphics[width=\columnwidth]{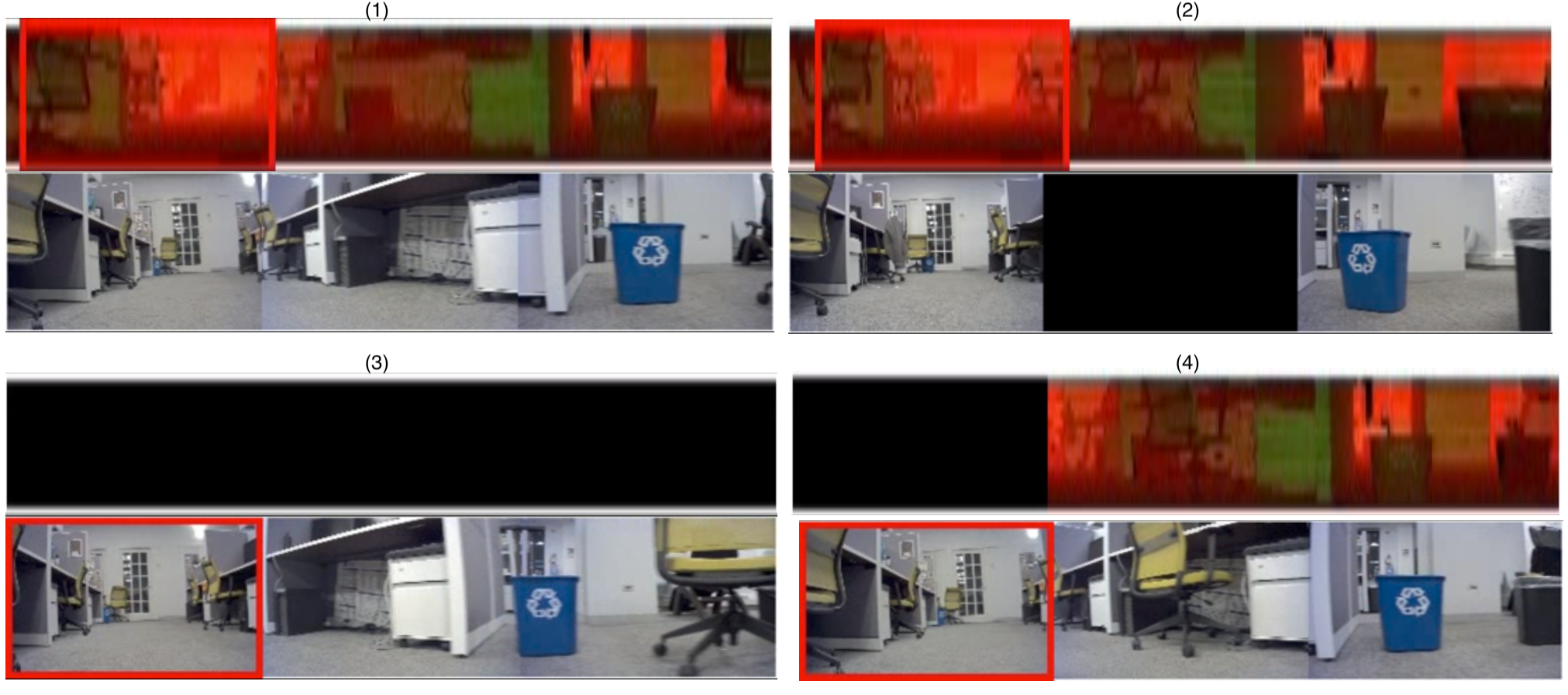}
    \caption{The snapshots of the video captured by four sensors when driving autonomously. The platform in each case was turning left with different inputs disabled. We present four cases: (1) all sensors are enabled, (2) the center camera is disabled, (3) the entire LiDAR sensor is disabled, (4) the left part of the LiDAR sensor is disabled. The red frame indicates the input selected by the Main Gating Network.}
    \label{fig:video}
    \vspace{-0.35in}
\end{figure}

In the \textcolor{yellow}{third} step of the training procedure, we train the final multi-modal experts network. We use the pre-trained experts from previous steps and set the learning rate to be $0.0001$. In particular, we use the pre-trained LiDAR expert obtained from the first step and pre-trained camera experts obtained in the second step in the case where LiDAR was disabled. Finally, we use all five settings of enabled/disabled sensors, to fine-tune the entire network in an end-to-end fashion. In particular, we lock the gating networks (will not update the parameters) and train the whole network. The test results of the obtained final network are shown in Fig.~\ref{5-branch-result} and Table~\ref{tbl:4s-final-result}, where the test performance is measured as the mean squared error(MSE) between the predicted and the reference steering command. The obtained network correctly chooses the useful sensor for driving and accurately predicts the steering command.
Finally, we compare the \textcolor{yellow}{test} performance of our model with several different approaches. The first three methods listed in Table~\ref{tbl:FLOPs} describe models that use inputs from single modality. They are constructed by one or more feature extractors followed by a fully-connected layer to make the final prediction. \textcolor{yellow}{The \textit{baseline} model is obtained by concatenating the features from all of the experts without any gating mechanism. It is trained end-to-end.} \textit{NetGated} and \textit{Sensor Dropout} were introduced in \cite{patel2017sensor} and \cite{Liu2017LearningEM} respectively. We followed the setups described in their original papers to train the networks. Also note that our approach is orthogonal to the existing network compression methods~\cite{ denton2014exploiting, gong2014compressing, han2015deep, han2015learning} since compression schemes are applicable to our approach and can lead to further computational reductions. To make fair comparisons, each network uses the same feature extractor architectures and the gating networks are designed to require significantly less computations than feature extractors, they are all described in the Appendix. We compare MSE between the predicted and actual steering command and the amount of computations, i.e. the number of floating-point operations (FLOPs) needed to perform a single inference. For the multisensory models, their performances are measured in five different scenarios when one of the sensors is disabled. The results are captured in Table~\ref{tbl:FLOPs}. The proposed network \textcolor{yellow}{achieves comparable performance to the \textit{baseline} model} and outperforms the majority of the competitive methods in terms of test loss except for \textit{NetGated}. However, \textcolor{yellow}{both the \textit{baseline} and \textit{NetGated} require significantly more computations and they do not have the ability to interpret the importance of each sensor at any given moment.} Meanwhile, our method uses almost the same amount of computations as the single modality networks. Therefore multi-modal experts network is superior over other sensor fusion techniques as it can make accurate predictions and handle sensor failures better while at the same time is computationally much more efficient.
We tested our proposed network on the autonomous platform and used the predicted steering angles to steer the vehicle in an indoor environment. The snapshots of the test video are shown in Fig.~\ref{fig:video}.

\begin{thisnote}

\subsection{Ablation Analysis}

\subsubsection{Can we use end-to-end training scheme to train the Multi-modal Experts Network?}

Note that the small gating network has its own feature extractors which should be correlated with the main feature extractors. Training our system with the end-to-end approach is not practical since it fails to learn these correlations, i.e. there is no mechanism to enforce correlation among mentioned feature extractors. Thus we propose a multi-step training method to first train the gating network and then fine-tune the experts based on the behaviors of the gating network. To further support this statement, we performed an experiment where we compare our performance with an end-to-end trained scheme (Fig.~\ref{fig:ablation}(a)) and demonstrate that simple end-to-end training approach does not work well for our setting.

\begin{figure}[t]
    \centering
    \vspace{0.1in}
    \includegraphics[width=0.95\columnwidth]{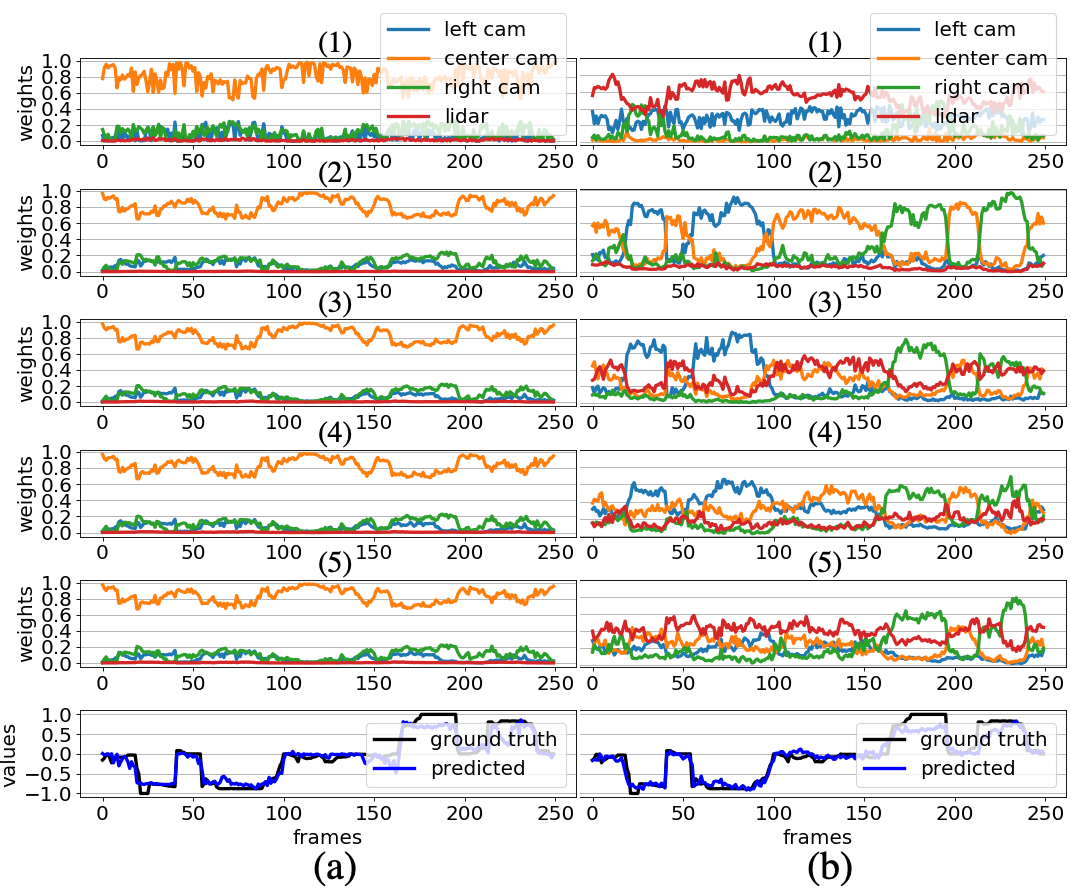}
    \caption{Ablation studies: the test results of individual gating network after 2.1 step of training when (a) The Main Gating Network is trained end-to-end together with the experts. (b) The gating mechanism is trained without the sparsity loss.}
    \label{fig:ablation}
    \vspace{-0.30in}
\end{figure}

\subsubsection{The effectiveness of $\mathcal{L}_{sparsity}$ and $\mathcal{L}_{nentropy}$}
To show the effectiveness of the two losses, we performed controlled experiments. We set both $\alpha$ and $\beta$ to zero when training the network with the gating mechanism in sub-steps 1.1 and 2.1. We show the test results obtained from the step 1.1 in Fig.~\ref{fig:lidar-gating}(b). When comparing this figure with Fig.~\ref{fig:lidar-gating}(a), we can see that without $\mathcal{L}_{sparsity}$, the network is less confident when selecting experts and without $\mathcal{L}_{nentropy}$ the selection is limited to a group of the following experts: left, center, and rightmost expert (see Fig.~\ref{fig:arch-5lidar} for explanation of the experts). This limitation therefore causes the LiDAR Gating Network to ignore the leftmost part of the LiDAR input. Furthermore, training without the sparsity loss (shown in Fig.~\ref{fig:ablation}(b)) demonstrate that the outputs of the gating network for each expert are closer to each other than when the loss is introduced (see Fig.~\ref{4s-final-result}). In other words, the Main Gating Network has less confidence in making the selection. The intuition is that without the sparsity loss, the experts tend to be cooperative instead of competitive with each other when predicting the steering angles.
\end{thisnote}
\section{CONCLUSION}
\label{sec:con}
This paper focuses on the autonomous driving problem with multiple sensing modalities, such as cameras and LiDAR. We consider the problem of efficient usage of multi-modal sensory resources with the goal of building a network that activates the processing of the data from only the relevant sensors. We show how to combine the discrete and continuous selection of relevant information. Our approach shows superiority over existing common baselines: it handles sensor failures, has high predictive power, and is efficient (runs in real time). In addition, by visualizing the output of the gating network, it becomes easy to see which part of the input is the most important for the driving task at any given moment. 






\section*{APPENDIX}
\vspace{-0.20in}
\begin{table}[H]
    \centering
  \caption{Network architecture}
  \vspace{-0.1in}
  \label{tbl:main-case6} 
    \begin{center}
LiDAR Expert:\\
The first 4 conv layers are followed by a 2x2 maxpooling with a stride of 2. An 1x2 maxpooling with a stride of (1,2) follows the conv5 layer.  \\
\vspace{0.02in}
\setlength{\tabcolsep}{5.2pt}
    \resizebox{0.75\columnwidth}{!}{
      \begin{tabular}{|c|c|c|} 
            \hline
            Layer name & output size &   Parameters \\
            \hline
            \hline
            conv1 & $16\times16\times148$ & k=$3\times5$, s=(1,1), p=(1,1), BN, ReLU \\
            \hline
            conv2 & $32\times8\times72$ & k=$3\times5$, s=(1,1), p=(1,1), BN, ReLU \\
            \hline
            conv3 & $64\times4\times34$ & k=$3\times5$, s=(1,1), p=(1,1), BN, ReLU \\
            \hline
            conv4 & $96\times2\times16$ & k=$3\times4$, s=(1,1), p=(1,1), BN, ReLU \\
            \hline
            conv5 & $128\times1\times8$ & k=$3\times3$, s=(1,1), p=(1,1), BN, ReLU \\
            \hline
            vectorize &  $512$ &\\
            \hline
      \end{tabular}}
    \end{center}
      \begin{center}
    \vspace{-0.04in}
      LiDAR Gating Network/  LiDAR feature extractor in Main Gating Network:\\
       A 2x2 maxpooling with a stride of 2 follows the conv2 and conv3 layer.  \\
\vspace{0.02in}
\setlength{\tabcolsep}{5.4pt}
\resizebox{0.75\columnwidth}{!}{
      \begin{tabular}{|c|c|c|} 
            \hline
            Layer name & output size & Parameters \\
            \hline
            \hline
            conv1 &$16\times10\times18$ &    k=$1\times1$, s=(2,27), p=(2,6), BN, ReLU \\
            \hline
            conv2 &$32\times8\times6$ &    k=$5\times7$, s=(1,3), p=(1,2), BN, ReLU \\
            \hline
            conv3 &$64\times4\times2$ &   k=$3\times4$, s=(1,1), p=(1,1), BN, ReLU \\
            \hline
            vectorize &$128$ &   \\
            \hline
      \end{tabular}}
    \end{center}
    \vspace{-0.04in}
    \begin{center}
    \vspace{-0.10in}    
    Camera Expert:\\
The conv1,2,4,5,6 layers are followed by a 2x2 maxpooling with a stride of 2. A 3x3 maxpooling with a stride of 2 follows the conv3 layer.  \\
\vspace{0.02in}
\setlength{\tabcolsep}{5pt}
\resizebox{0.75\columnwidth}{!}{
      \begin{tabular}{|c|c|c|} 
            \hline
            Layer name & output size &  Parameters \\
            \hline
            \hline
            conv1 &$16\times60\times96$ & k=$4\times4$, s=(2,2), p=(1,1), BN, ReLU\\
            \hline
            conv2 &$32\times30\times48$ &    k=$3\times3$, s=(1,1), p=(1,1), BN, ReLU \\
            \hline
            conv3 &$64\times16\times24$ &  k=$2\times3$, s=(1,1), p=(1,1), BN, ReLU\\
            \hline
            conv4 &$96\times8\times12$ &   k=$3\times3$, s=(1,1), p=(1,1), BN, ReLU \\
            \hline
            conv5 &$128\times4\times6$ &   k=$3\times3$, s=(1,1), p=(1,1), BN, ReLU \\
            \hline
            conv6 &$256\times2\times4$ &   k=$3\times2$, s=(1,1), p=(1,1), BN, ReLU \\
            \hline
            vectorize &$512$ &   \\
            \hline
      \end{tabular}}
    \end{center}
      \begin{center}
    \vspace{-0.04in}
                Camera feature extractor in Main Gating Network:\\
The conv2,3 layers are followed by a 2x2 maxpooling with a stride of 2.
\vspace{0.02in}
\setlength{\tabcolsep}{4pt}
\resizebox{0.75\columnwidth}{!}{
      \begin{tabular}{|c|c|c|} 
            \hline
            Layer name & output size & Parameters \\
            \hline
            \hline
            conv1 &$16\times13\times20$ &    k=$1\times1$, s=(10,10), p=(1,1), BN, ReLU\\
            \hline
            conv2 &$32\times6\times10$ &    k=$5\times4$, s=(2,2), p=(1,1), BN, ReLU \\
            \hline
            conv3 &$64\times2\times4$ &   k=$4\times4$, s=(1,1), p=(1,1), BN, ReLU \\
            \hline
            vectorize &$128$ &   \\
            \hline
      \end{tabular}}
    \end{center}

            \vspace{-0.02in}
            LiDAR\_full expert in LiDAR only network:\\
The first four conv layers are followed by a 2x2 maxpooling with a stride of 2. An 1x2 maxpooling with a stride of (1,2) follows the conv5 layer.  \\
\vspace{0.02in}
\setlength{\tabcolsep}{5pt}
\resizebox{0.75\columnwidth}{!}{
      \begin{tabular}{|c|c|c|} 
            \hline
            Layer name & output size & Parameters \\
            \hline
            \hline
            conv1 &$16\times16\times222$ &    k=$3\times10$, s=(1,2), p=(1,1), BN, ReLU \\
            \hline
            conv2 &$32\times8\times106$ &    k=$3\times8$, s=(1,1), p=(1,1), BN, ReLU \\
            \hline
            conv3 &$64\times4\times50$ &   k=$3\times6$, s=(1,1), p=(1,1), BN, ReLU \\
            \hline
            conv4 &$96\times2\times24$ &   k=$3\times4$, s=(1,1), p=(1,1), BN, ReLU \\
            \hline
            conv5 &$128\times1\times12$ &   k=$3\times3$, s=(1,1), p=(1,1), BN, ReLU \\
            \hline
            linear & $512$ & $768\times512$  \\
            \hline
      \end{tabular}}
      \vspace{-0.1in}
\end{table}

\section*{ACKNOWLEDGMENT}
We acknowledge the support of NVIDIA Corporation with the donation of the Drive PX2 used for this research.


\enlargethispage{12cm}
\IEEEtriggeratref{7}
\bibliographystyle{IEEEtran}
\bibliography{IEEEexample}

\end{document}